\documentclass[10pt,twocolumn,letterpaper]{article}

\usepackage{iccv}
\usepackage{times}
\usepackage{epsfig}
\usepackage{graphicx}
\usepackage{amsmath}
\usepackage{amssymb}
\usepackage{capt-of}
\usepackage{caption}
\usepackage{booktabs}
\usepackage{dashrule}
\usepackage{cuted}
\usepackage[square,comma,sort&compress,numbers]{natbib}

\usepackage[pagebackref=true,breaklinks=true,letterpaper=true,colorlinks,bookmarks=false]{hyperref}

\usepackage{multirow}
\usepackage{amssymb}
\usepackage{tikz}

\newcommand{\tikzcircle}[2][red,fill=red]{\tikz[baseline=-0.5ex]\draw[#1,radius=#2] (0,0) circle ;}%

\graphicspath{{figures/}} 

\iccvfinalcopy 

\def\iccvPaperID{xxxx} 

\ificcvfinal\fi

\begin{document}
	
\title{AniPortraitGAN: Animatable 3D Portrait Generation from 2D Image Collections\!}

\author{Yue Wu$^{1}$\thanks{Equal contribution. Work done when YW was an intern at MSRA.} \quad Sicheng Xu$^{2\ \!*}$ \quad Jianfeng Xiang$^{3,2}$ \quad Fangyun Wei $^{2}$ \\ Qifeng Chen$^{1}$ \quad Jiaolong Yang$^{2}$\thanks{Corresponding author and project lead.} \quad Xin Tong$^{2}$ \\
	$^1${HKUST} \quad  $^2${Microsoft Research Asia} \quad $^3${Tsinghua University}
}

\maketitle
\ificcvfinal\thispagestyle{empty}\fi

\begin{strip}
	\vspace{-40pt}
	\centering
	\includegraphics[width=1\linewidth]{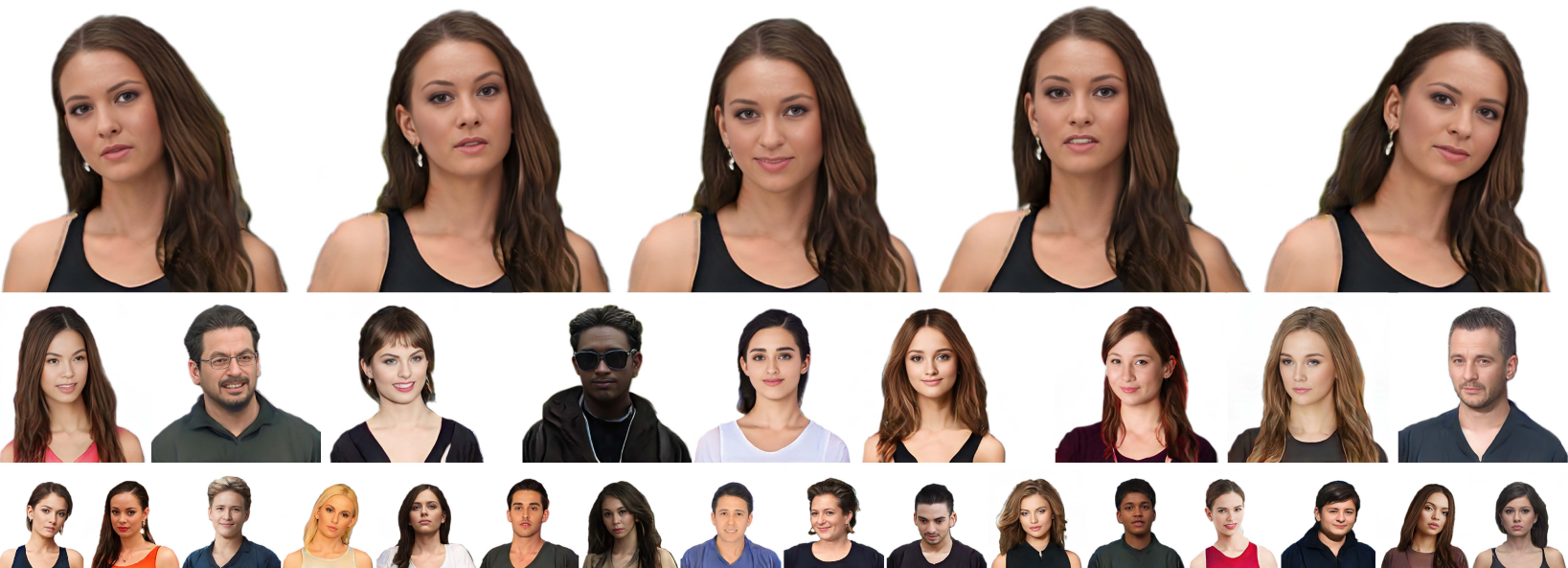}
	\vspace{-16pt}
	\captionsetup{type=figure,font=small}
	\caption{Our method is a new 3D-aware GAN that can generate diverse \emph{virtual} human portraits (512$\times$512) with explicitly controllable 3D camera viewpoints, facial expression, head pose, and shoulder movements. It is trained on unstructured 2D images without any 3D or video data. (\textbf{Best viewed with zoom; see our \emph{\href{https://yuewuhkust.github.io/AniPortraitGAN/}{project page}} for videos of more samples})} 
	\label{fig:teaser}
	\vspace{8pt}
\end{strip}

\begin{abstract}
	Previous animatable 3D-aware GANs for human generation have primarily focused on either the human head or full body. However, head-only videos are relatively uncommon in real life, and full body generation typically does not deal with facial expression control and still has challenges in generating high-quality results. Towards applicable video avatars, we present an animatable 3D-aware GAN that generates portrait images with controllable facial expression, head pose, and shoulder movements.
	It is a generative model trained on unstructured 2D image collections without using 3D or video data. 
	For the new task, we base our method on the generative radiance manifold representation and equip it with learnable facial and head-shoulder deformations. 
	A dual-camera rendering and adversarial learning scheme is proposed to improve the quality of the generated faces, which is critical for portrait images. A pose deformation processing network is developed to generate plausible deformations for challenging regions such as long hair. Experiments show that our method, trained on unstructured 2D images, can generate diverse and high-quality 3D portraits with desired control over different properties. 
\end{abstract}

\section{Introduction}

The automatic creation of animatable 3D human characters has become an increasingly important topic  with a range of applications including video conferencing, movie production, and gaming. The related techniques have undergone a significant growth recently, with a variety of promising methods being proposed~\cite{deng2020disentangled,wu2022anifacegan,sun2022controllable,gafni2021dynamic,bergman2022generative,wang2022morf,chen2022gdna,sun2023next3d,noguchi2022unsupervised,peng2021animatable,ouyang2022real,hong2023eva3d,xu2023omniavatar}.

Among these techniques, 3D-aware generation methods have emerged as a particularly promising avenue~\cite{wu2022anifacegan,sun2022controllable,bergman2022generative,noguchi2022unsupervised,sun2023next3d,xu2023omniavatar,hong2023eva3d,dong2023ag3d}. These methods can leverage the ample availability of unstructured 2D data for 3D human generative learning without the need for 3D scans or multiview human images which are difficult to acquire at scale. Typically, these methods employ Generative Adversarial Networks (GANs)~\cite{goodfellow2014generative} for unsupervised training, use neural implicit fields~\cite{mildenhall2020nerf} as the 3D representation, and incorporate priors from 3D face and body parametric models~\cite{blanz1999morphable,li2017learning,loper2015smpl} for character control.

Despite their promising potential, existing animatable 3D-aware human generation focuses on either the human head or full body and have encountered limitations in their applicability. Head generation methods~\cite{wu2022anifacegan,sun2022controllable,sun2023next3d,xu2023omniavatar} can produce high-quality face and hair with controllable facial expression. Unfortunately, videos featuring only a human head are relatively uncommon in everyday life, and therefore these methods are less applicable in practical scenarios. Full body generation~\cite{bergman2022generative,noguchi2022unsupervised,zhang2022avatargen,hong2023eva3d,dong2023ag3d}, on the other hand, also generates torso and limbs with explicit pose control. However, generating high-quality full body human is still challenging due to the complexity of body motion. In addition, the facial region is often underrepresented in these full body methods and there is no expression control.

Our paper presents a new 3D-aware generation method that is the first to focus on animatable generation of the human head and shoulder regions. Our method enables fine-grained control over facial expressions as well as head and shoulder movements, making it well suited for real-world applications such as video conferencing and virtual presenters. Like previous 3D-aware GANs, our method is trained on unstructured 2D image sets.

For this new task, we follow previous methods to train neural radiance generation with 3D parametric model priors in a GAN training scheme. 
We base our method on the 3D-aware GAN framework of GRAM~\cite{deng2022gram,xiang2022gram} and follow AniFaceGAN~\cite{wu2022anifacegan} for facial expression control using 3D morphable model (3DMMs) priors. For head and shoulder control, we incorporate the SMPL~\cite{loper2015smpl} body model for deformation guidance. 
However, we found that naively extending these existing techniques to our animatable head-shoulder portrait generation task is deficient. One prominent issue is about face quality, which is of paramount importance for visual communications. To handle the complex image distribution caused by the large variations of head position and orientation, we propose a dual-camera rendering and adversarial learning scheme for training. An additional dynamic camera is placed around human head pointing at head center to render faces for discrimination, which significantly improves the face generation quality. 
Another issue is the SMPL-guided human body deformation, for which we identified that the commonly-used strategy based on linear blending skinning failed to generate convincing results for human characters with long hair. Sharp discontinuities will occur under head rotation for hair region, leading to significant artifacts. To tackle this issue, we propose a pose deformation volume processing module to learn better deformations, which stabilizes GAN training and produces visually plausible results. 

We train our model on a head-shoulder portrait dataset called SHHQ-HS, which is constructed by cropping and superresolving the 40K human body images in the SHHQ dataset \cite{fu2022stylegan}. We show that our method can generate diverse and high-quality 3D portrait images with flexible control of different properties including facial expressions and head-shoulder poses.

Our contributions can be summarized as follows:
\begin{itemize}
	\item We propose the first animatable 3D-aware portrait GAN that generates head and shoulder regions with facial expression and head-shoulder motion control. We believe generating such animatable human characters is a missing piece of 3D-aware human GANs for real-world applications like video conferencing and virtual presenters.
	\item We propose a dual-camera rendering and adversarial learning scheme that gives rise to high-quality face generation comparable to previous head-only 3D-aware GANs.
	\item We propose a pose deformation processing module which achieves smooth and plausible pose-driven deformation for human hair.
\end{itemize}

\section{Related Work}

\paragraph{3D-aware Image Generation} 3D-aware image generative models aim to generate images that allow for the explicit control of 3D camera viewpoint, training only on 2D images. Most existing works are based on the GAN framework for generative modeling. Early approaches to this problem use 3D convolutions to generate 3D feature volumes and project them to 2D plane for image generation~\cite{nguyen2019hologan,nguyen2020blockgan}. Recently, methods based on more explicit 3D representations and differentiable rendering have become popular \cite{liao2020towards,schwarz2020graf,niemeyer2021giraffe,shi2021lifting,chan2021pi,devries2021unconstrained,gu2021stylenerf,chan2022efficient,deng2022gram,xiang2022gram,zhao2022generative,skorokhodov2022epigraf,or2022stylesdf,tewari2022disentangled3d,skorokhodov20233d}. The widely-used 3D representations are NeRF~\cite{mildenhall2020nerf} and its variants, for their strong capability to model real-world 3D scenes. 

Among these NeRF-based GANs, some use volume rendering to directly generate the final images, which ensures strong 3D consistency among different views but often has high computation cost. Others apply 2D convolutions on the rendered low-resolution 2D images or feature maps for upsampling, which significantly reduces the computation cost but sacrifices multiview consistency for the generated instances. Our method uses the high-resolution radiance manifold representation of \cite{xiang2022gram}, which can generate high resolution images with strong multiview consistency.

\paragraph{Controllable Human Head and Body Generation}
Adding explicit controls to face and body generative modeling has received much attention in recent years. Existing works have primarily focused on either the
head~\cite{deng2020disentangled,wu2022anifacegan,sun2022controllable,bergman2022generative,sun2023next3d,xu2023omniavatar} or the whole body \cite{bergman2022generative,chen2022gdna,noguchi2022unsupervised,hong2023eva3d,dong2023ag3d}, with priors from 3D parametric models being commonly incorporated to achieve semantically meaningful control. For expression control, most head GANs \cite{deng2020disentangled,wu2022anifacegan,sun2022controllable,bergman2022generative,sun2023next3d} often incorporate 3D morphable models (3DMMs)~\cite{blanz1999morphable} or FLAME models~\cite{li2017learning} in their training process. Whole-body GANs typically rely on the SMPL model~\cite{loper2015smpl} for body pose animation (with a few exceptions such as \cite{noguchi2022unsupervised} that use body skeleton), and they often do not address facial expression control. This work deals with a new human generation task: generating portrait figures that contain head and shoulder, with controllable facial expression and head-shoulder poses.

\paragraph{Human Image and Video Manipulation}
Our method is also related to human image and video manipulation approaches~\cite{wiles2018x2face,siarohin2019first,zakharov2019few,xu2020deep,wang2021one,gafni2021dynamic,liu2021neural,ren2021pirenderer,doukas2021headgan,peng2021animatable,jiang2022nerffaceediting,kwon2021neural,gao2022mps,weng2022humannerf} that also produce human animation videos. However, the goal and underlying techniques of these methods differ significantly from ours. These methods aim to animate the human character in the given image or video, and are typically trained in a supervised manner using videos or image pairs. In contrast, we deal with human generative modeling and novel character creation, training on unstructured still images in an unsupervised or weakly-supervised fashion.

\begin{figure*}[t!]
	\centering
	\includegraphics[width=\textwidth]{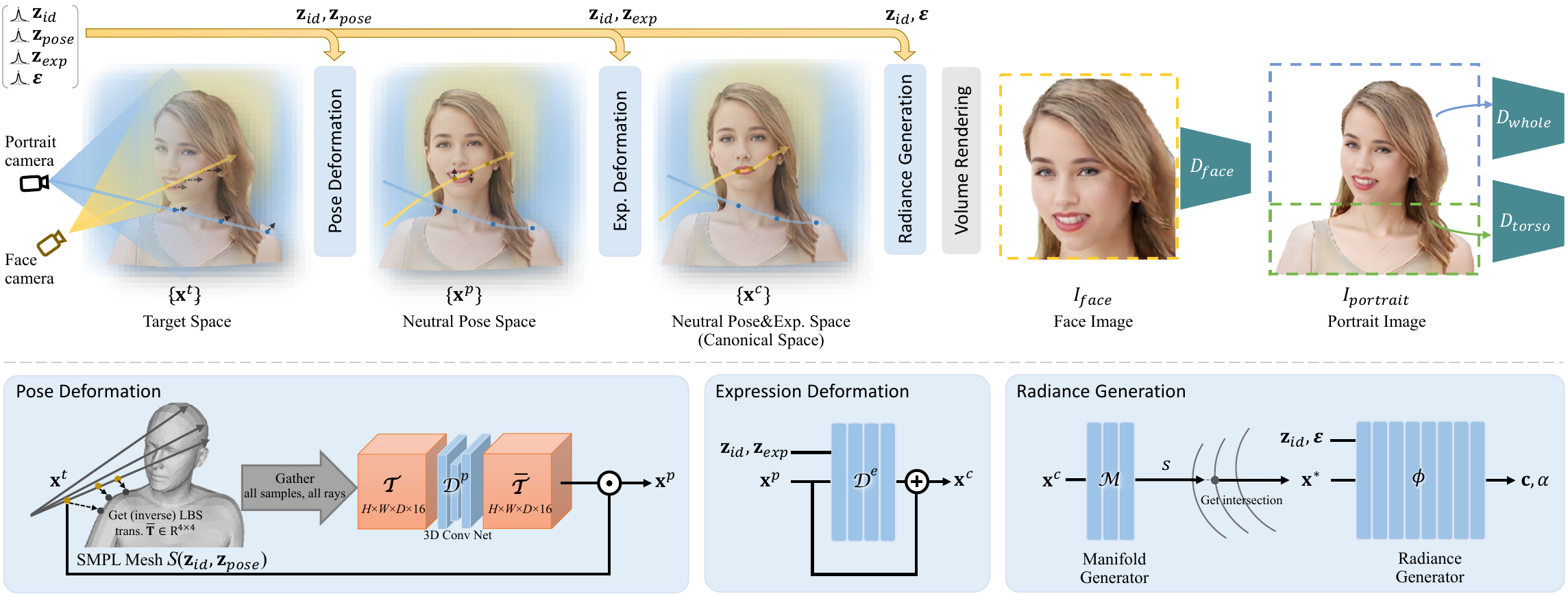}
	\vspace{-17pt}
	\caption{Method overview. \emph{Top}:  the pipeline of our controllable 3D-aware portrait GAN. For training, we apply a dual-camera rendering scheme with two images separated rendered and three discriminators employed. \emph{Bottom}: structures of the deformation and radiance generation modules (the radiance manifold super-resolution step is omitted for simplicity; see text for details). 
	}
	\label{fig:framework}
\end{figure*}

\section{Method}

Our goal is to generate human portrait images containing human head and shoulder regions by training on a given 2D image collection. As in a standard GAN setup, we sample random latent codes and map them to the final output image.  The input to our generator consists of multiple latent codes, which corresponds to different properties of the generated human, and the camera viewpoint. The output is a human portrait image carrying the desired properties.

Figure \ref{fig:framework} presents an overview of our method. The overall pipeline follows the popular paradigm of canonical neural radiance representation in combination with (inverse) deformation ~\cite{wu2022anifacegan,bergman2022generative,hong2023eva3d}. 

\subsection{Latent Codes}\label{sec:codes}
Our latent codes contain an identity code $\mathbf{z}_{id}\!\in\!\mathbb{R}^{d_i}$ for human shape, an expression code $\mathbf{z}_{exp}\!\in\!\mathbb{R}^{d_e}$ for facial expression, an $\mathbf{z}_{pose}\!\in\!\mathbb{R}^{d_p}$ for head and shoulder pose, and an additional noise $\boldsymbol{\varepsilon}\in\mathbb{R}^{d_\varepsilon}$ controlling other attributes such as appearance.
To achieve semantically meaningful control, we incorporate priors 3D human parametric models and align our latent space with theirs.
Specifically, our identity code $\mathbf{z}_{id}$ is designed as the concatenation of 3DMM~\cite{paysan20093d} face identity coefficient and  SMPL~\cite{loper2015smpl} body shape coefficient. The pose code $\mathbf{z}_{pose}$ is a reduced SMPL pose parameter, which consists of the joint transformations of 6 joints: head, neck, left and right collars, and left and right shoulders. The expression code $\mathbf{z}_{exp}$ is the same as 3DMM expression coefficient.

\subsection{Canonical Radiance Manifolds} \label{sec:radiance}
Our method utilizes the radiance manifolds \cite{deng2022gram,xiang2022gram} to represent canonical humans.
This representation regulates radiance field learning and rendering on a set of learned implicit surfaces in the 3D volume. It can generate high-quality human faces with strict multiview consistency. With manifold superresolution~\cite{xiang2022gram}, it can generate high-resolution images efficiently without sacrificing multiview consistency.

Concretely, we apply three networks for radiance generation. A manifold prediction MLP $\mathcal{M}$ takes a point $\mathbf{x}$ in the canonical space as input and predicts a scalar $s$:
\begin{equation}
	\mathcal{M}:\mathbf{x}\in\mathbb{R}^3\rightarrow s\in\mathbb{R}.
\end{equation}
It models a scalar field that defines the surfaces. 
A radiance generation MLP $\phi$ generates the color and opacity for points on the surfaces given the identity codes $\mathbf{z}_{id}$, noise $\boldsymbol{\varepsilon}$ and view direction $\mathbf{d}$:
\begin{equation}
	\mathcal{\phi}:(\mathbf{x},\mathbf{z}_{id},\boldsymbol{\varepsilon},\mathbf{d})\in\mathbb{R}^{d_i+d_\varepsilon+6}\rightarrow (\mathbf{c},\alpha)\in\mathbb{R}^4.
\end{equation}
A manifold superresolution CNN $\mathcal{U}$ upsamples the flattened and discretized radiance maps $\mathbf{R}_{lr}$ to high-resolution ones $\mathbf{R}_{hr}$:
\begin{equation}
	\mathcal{U}:(\mathbf{z}_{id},\boldsymbol{\varepsilon},\mathbf{R}_{lr})\rightarrow \mathbf{R}_{hr},
\end{equation}
for which we use a $128^2\rightarrow512^2$ upsampling setting in this paper.
For more technical details, we refer the readers to \cite{deng2022gram,xiang2022gram}.

\subsection{Deformation Fields}\label{sec:deform}

For each sampled 3D point in the target space with desired head-shoulder pose and facial expression, we apply deformations to transform them to the canonical space for radiance retrieval. A two-stage deformation scheme is used to neutralize pose and expression.

\subsubsection{Pose Deformation Generator} 

We incorporate the SMPL model \cite{loper2015smpl} and use its linear blend skinning (LBS) scheme~\cite{lewis2000pose} to guide our deformation. Given the shape code $\mathbf{z}_{id}$ and pose code $\mathbf{z}_{pose}$, a posed human body mesh can be constructed using SMPL. 
The SMPL model provides a pre-defined skinning weight vector $\mathbf{w}\in\mathbb{R}^{N_J}$ for each vertex on the body surface, where $N_J$ is the joint number. 

A simple approach for propagating body surface deformation to the full 3D space is assigning any point the skinning weights of its closest body surface vertex and use them to deform it. In fact, this strategy is widely used in state-of-the-art animatable human body modeling and generation methods~\cite{huang2020arch,peng2021animatable,bhatnagar2020loopreg,gao2022mps,zhang2022avatargen,dong2023ag3d}. While it yields plausible results for existing full-body synthesis method, we find it incurs significant visual defects in high-resolution portrait synthesis. For human characters with long hairs, this native strategy leads to sharp deformation discontinuity for the hair regions above shoulders (see Fig.~\ref{fig:ablation_dualcam_lbs}).

We propose a deformation volume processing module to tackle this issue.
Specifically, for a point $\mathbf{x}^t$ in the target space with the skinning weight vector $\mathbf{w}$ retrieved from the closest SMPL body vertex, the deformed point can be calculated using inverse LBS:
\begin{equation}\textstyle 
	\mathbf{x}^p=LBS^{-1}(\mathbf{x}^t,{\mathbf{w}})=\bar{\mathbf{T}}\!\cdot\!\mathbf{x}^t=\big(\sum_{j=1}^{N_J}{w}_j\mathbf{T}_j\big)\!\cdot\!\mathbf{x}^t,
\end{equation}
where $\bar{\mathbf{T}}\in\mathbb{R}^{4\times 4}$ is a transformation matrix computed by linearly blending the SMPL joint transformations $\mathbf{T}_j\in\mathrm{SE(3)}$.
We gather the reshaped transformation matrices for the sampled points of all $H\times W$ rays into a tensor $\boldsymbol{\mathcal{T}}\in\mathbb{R}^{H\times W \times D \times 16}$, where $D$ is the number of sampled points per ray, and apply a 3D CNN $\mathcal{D}^p$ to process it:
\begin{equation}
	\mathcal{D}^p:\boldsymbol{\mathcal{T}}\in\mathbb{R}^{H\times W \times D \times 16}\rightarrow \overline{\boldsymbol{\mathcal{T}}}\in\mathbb{R}^{H\times W \times D \times 16}.
\end{equation}
After processing, we reshape the transformations back and apply them to the sampled points to accomplish pose deformation.

\subsubsection{Expression Deformation Generator} We further apply a deformation to neutralize the facial expression from the target expression. Following \cite{wu2022anifacegan}, we introduce a deformation field which is guided by the 3DMM model~\cite{paysan20093d}. Specifically, an MLP $\mathcal{D}^e$ is employed to deform the points in the pose-aligned space:
\begin{equation}
	\mathcal{D}^e:(\mathbf{x}^p,\mathbf{z}_{id}, \mathbf{z}_{exp})\rightarrow \mathbf{x}^c.
\end{equation}
This deformation network will be trained to generate faces with expressions following 3DMM:
\begin{equation}
	\mathbf{S} = \mathbf{S}(\mathbf{z}_{id},\mathbf{z}_{exp}) = \bar{\mathbf{S}}+\mathbf{B}_{id}\mathbf{z}_{id} + \mathbf{B}_{exp}\mathbf{z}_{exp},
\end{equation}
where $\bar{\mathbf{S}}$ is the 3DMM mean face, and $\mathbf{B}_{id}$ and $\mathbf{B}_{exp}$ are the PCA basis for identity and expression, respectively. Training details can be found in later sections.

In total, our generator $G$ has five sub-nets: $\mathcal{M}$, $\phi$, $\mathcal{U}$, $\mathcal{D}^p$, and $\mathcal{D}^e$. For final image rendering,  we calculate $M$ intersection points $\{\mathbf{x}^{*}_i\}$ between a deformed ray $\mathbf{r}$  (point samples) 
and the canonical manifolds. We then obtain the color and occupancy of $\{\mathbf{x}^*_i\}$ by sampling the radiance map $\mathbf{R}_{hr}$, and composite the color via:
\begin{equation}
	C({\mathbf{r}}) = \sum_{i=1}^{M}T({\mathbf{x}}^*_i)\alpha(\mathbf{x}^*_i)\mathbf{c}(\mathbf{x}^*_j) 
	,\ \  T({\mathbf{x}}^*_i) = \prod_{k<i}(1-\alpha(\mathbf{x}^*_k)). \label{eq:render}
\end{equation}

\subsection{Dual-Camera Discriminators}
Previous 3D-aware head GANs have demonstrated striking face generation quality by carefully center-aligning the generated and real face images for training. In our case, however, the head region constitutes part of the portrait image and its spatial position and orientation vary significantly. Simply applying a whole-image discriminator cannot offer adequate supervision for high-quality face generation, which is crucial for portrait images.

A straightforward remedy is to crop and align the faces in the rendered images and apply a local face discriminator. 
However, since image resampling operators are inherently low-pass, such an image-space cropping strategy introduces blur to the cropped faces, which is detrimental to GAN training especially for a multi-discriminator setup.
In this work, we design a dual-camera rendering scheme for GAN training. In addition to the main camera for full portrait image rendering,  we add another camera for face rendering, which is placed around human head pointing at head center, as shown in Fig.~\ref{fig:framework}. It is designed to have the same local coordinate system as in previous 3D-aware head GANs~\cite{deng2022gram,wu2022anifacegan}, and its position can be readily computed using the deformed SMPL head. Another possible idea is blending the output of two separate generators for face and body, as in some 2D human generation methods~\cite{fruhstuck2022insetgan}. But applying this strategy to the 3D, animatable case seems not straightforward.

Adding a dedicated face camera for training not only avoids image resampling and provides a more direct supervision to the canonical radiance manifolds, but also enables higher-resolution face rendering for adversarial learning. 
Consequently, the radiance generator receives stronger supervision for the face region. We apply two image discriminators $D_{whole}$ and $D_{face}$ for the rendered portrait image and face image with these two cameras, respectively. We also add another local discriminator $D_{torso}$, which takes the lower $1/4$ part of the rendered portrait images as input.

\subsection{Training Losses}

\paragraph{Adversarial Learning.} 
We apply the non-saturating GAN loss with R1 regularization \cite{mescheder2018training} for the 3D-aware image generator and all three discriminators $D_{whole}$, $D_{face}$, and $D_{torso}$. We empirically set the balancing weights to be $\lambda_{whole}= 0.1$,  $\lambda_{face}= 1.0$ and $D_{torso}=0.5$, respectively.

\paragraph{Deformation Learning} 
Following \cite{wu2022anifacegan}, we use a 3D landmark loss and imitation loss  
to gain expression control with 3DMM guidance. The landmark loss enforces the generated face image to have similar 3D facial landmarks to the 3DMM face constructed with the input identity and expression codes:
\begin{equation}
	\mathcal{L}_{lm} = \left \|f_{lm}(\mathbf{S}(\mathbf{z}_{id}, \mathbf{z}_{exp})), f_{lm}(\mathbf{S}(\hat{\mathbf{z}}_{id}, \hat{\mathbf{z}}_{exp}))\right \|_2^2, \label{eq:loss_lm}
\end{equation}
where $\hat{\mathbf{z}}_{id}$, $\hat{\mathbf{z}}_{exp}$ are the 3DMM coefficients estimated from the generated image using a face reconstruction network~\cite{deng2019accurate} and $f_{lm}$ denotes a simple facial landmark extraction function. 
For deformation imitation, we enforce the displacement of an input point $\mathbf{x}^p$ to follow its nearest point $\mathbf{x}^p_{ref}$ on the 3DMM mesh:
\begin{equation}
	\mathcal{L}_{{\rm 3DMM}} = \left \| \big(\mathcal{D}^e(\mathbf{x}^p)-\mathbf{x}^p\big) - \Delta \mathbf{x}^p_{ref}  \right \|_2^2 , \label{eq:3dmm}
\end{equation}
where $\Delta \mathbf{x}^p_{ref} = -\mathbf{B}_{exp}\mathbf{z}_{exp}(\mathbf{x}^p_{ref})$ is the 3DMM-derived (inverse) deformation of point $\mathbf{x}^p_{ref}$.

\begin{figure*}[t!]
	\centering
	\includegraphics[width=\textwidth]{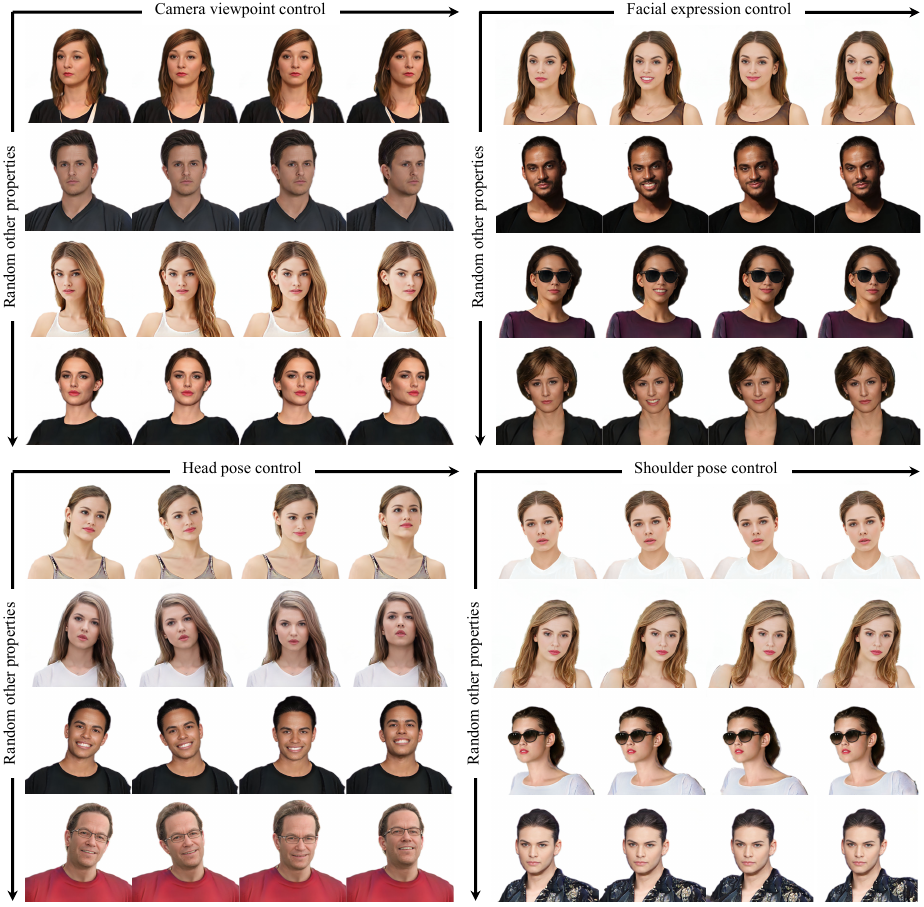}
	\vspace{-18pt}
	\caption{Results with controllable camera viewpoint, facial expression, head pose and shoulder pose. (\textbf{Best viewed with zoom})}
	\label{fig:results_control}
\end{figure*}

We impose several regularizations on the deformations. First, we encourage the deformations to be as smooth as possible. For pose deformation processing, we apply
\begin{equation}
	\mathcal{L}_{pose\_smooth} = \left \| \mathcal{D}^p(\boldsymbol{\mathcal{T}}) - {AvgPool}(\boldsymbol{\mathcal{T}})\right \|_2^2. \label{eq:pose_deform_smooth}
\end{equation}
where $AvgPool$ is a fixed average pooling operator. For the expression deformation, we apply
\begin{equation}
	\mathcal{L}_{exp\_smooth} = \left \| \mathcal{D}^e(\mathbf{x}^p) - \mathcal{D}^e(\mathbf{x}^p+\Delta) \right\|_2^2,  
\end{equation}
where $\Delta$ is a small random perturbation. Finally, a minimal deformation
constraint is applied for the expression deformation:
\begin{equation}
	\mathcal{L}_{exp\_minimal} = \left \| \mathbf{x}^p - \mathcal{D}^e(\mathbf{x}^p)\right \|_2^2. \label{eq:reg}
\end{equation}

\subsection{Training Strategy}

We employ a two-stage training strategy to train our model. At the first stage, we train a low-resolution image generator and the corresponding discriminators. Both the face and portrait branches generate $128\times 128$ images. All sub-networks are trained except for the manifold superresolution CNN $\mathcal{U}$. For the second stage, we generate $512\times 512$ portrait images and $256\times 256$ faces. We randomly initialize and train $\mathcal{U}$ as well as the high-resolution discriminators with all other sub-networks frozen.

\section{Experiments}
\paragraph{Training Data} We build a training set by processing the human images in the SHHQ dataset~\cite{fu2022stylegan}. SHHQ contains 40K full-body images of $1024\times 512$ resolution. To obtain high-quality head-shoulder portraits, we first fit SMPL models on the SHHQ images using the method of \cite{joo2021exemplar}.
Then, we crop the images and align them using the projected head and neck joints. The cropped portrait images are about $256\times 256$ resolution. We upsample them using the super-resolution methods of \cite{wang2021towards} and \cite{wang2018esrgan} to $1024\times 1024$, followed by downsampling them to $512\times 512$. Finally, the backgrounds are removed by applying the provided segmentation masks. We call this dataset \emph{SHHQ-HS}.

\paragraph{Implementation Details}
Our manifold predictor and radiance generator follow the implementations of \cite{deng2022gram}. 24 radiance manifolds are used as in \cite{deng2022gram}. The manifold superresolution net $\mathcal{U}$ is a smaller CNN compared to that in \cite{xiang2022gram}.
The pose deformation CNN $\mathcal{D}^p$ has two 3D conv layers with kernel size $9\!\times\!9\times\!5$. The expression deformation network $\mathcal{D}^e$ is the same as \cite{wu2022anifacegan}. 
See Fig.~\ref{fig:network_structure} for more details. 
For all experiments, we use the Adam optimizer~\cite{kingma2015adam} for training. 
Prior to training, we estimate the identity, pose and expression coefficients as well as camera poses for the images in the dataset using 3DMM and SMPL fitting \cite{deng2019accurate,joo2021exemplar}. 
During training, we randomly sample latent codes $\mathbf{z}_{id}$, $\mathbf{z}_{pose}$, and $\mathbf{z}_{exp}$ and camera pose $\boldsymbol{\theta}$ from the estimated distributions, and sample $\boldsymbol{\varepsilon}$ from a normal distribution. 

\paragraph{Runtime.} With an unoptmized implementation, our method takes about 0.87 seconds to generate one $512\times 512$ image from a set of given latent codes, evaluated on a NVIDIA RTX A6000 GPU.

\subsection{Generation Results}
Figure~\ref{fig:teaser} 
and \ref{fig:results_uncurated} 
present some generated portraits from our method trained on SHHQ-HS. The results are diverse and of high-quality, with camera viewpoint, facial expression, head rotation, and shoulder pose explicitly controlled. 
Figure~\ref{fig:results_control} shows the generated results for which we control one property out of the four while randomly changing the others.
Our method achieves consistent control for all the four properties for different identities. More results can be found in the suppl. video.

\subsection{Comparison with Previous Methods}

We compare our method with three state-of-the-art 3D-aware GANs: EG3D~\cite{chan2022efficient}, GRAM-HD~\cite{xiang2022gram} and AniFaceGAN~\cite{wu2022anifacegan}. Note that to our knowledge, there is no previous work that deals with the animatable head-shoulder portrait generation task in this paper, and hence we compare with these three methods for reference purpose only.

\begin{figure}[t!]
	\includegraphics[width=\columnwidth]{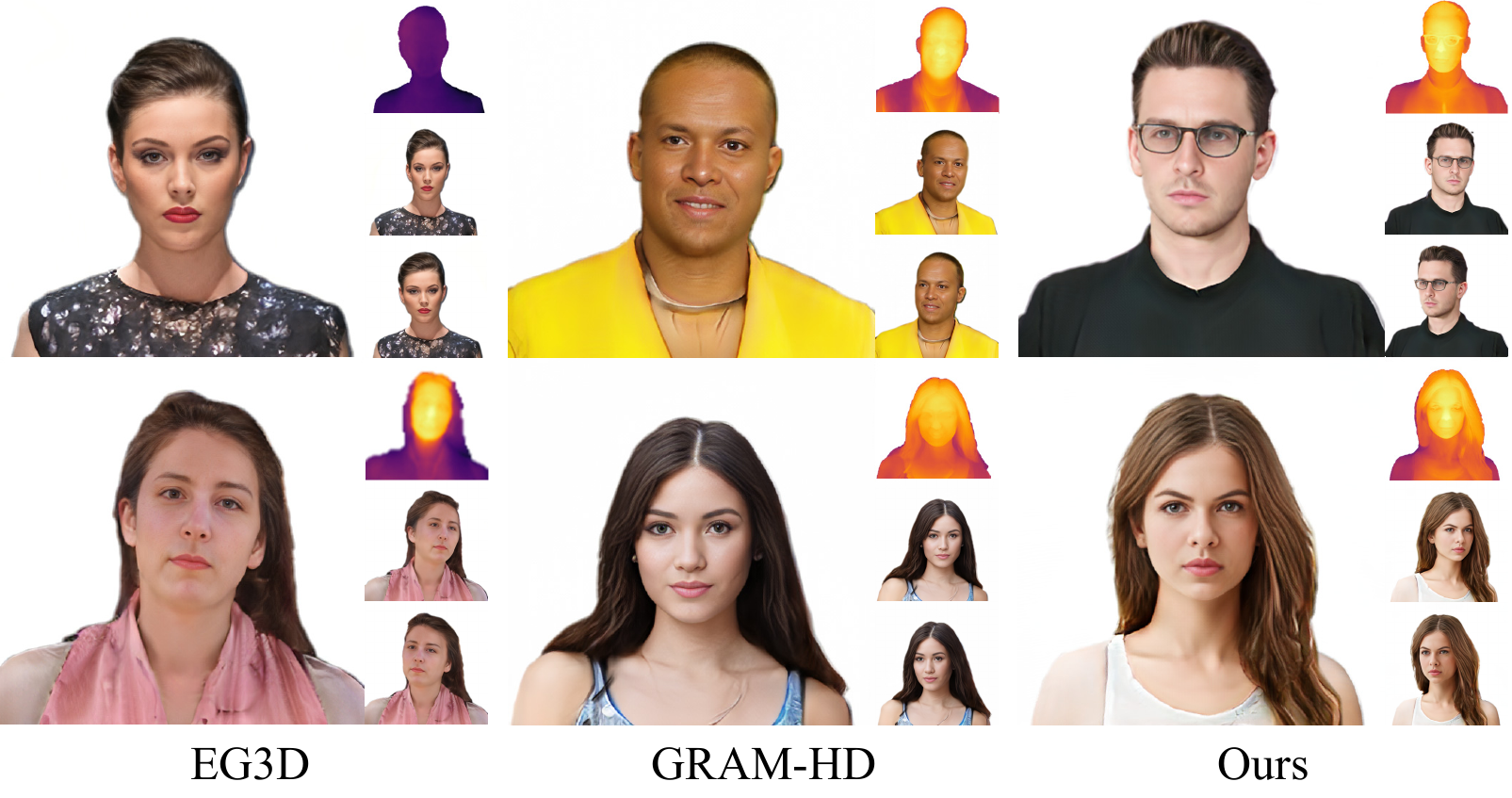}
	\vspace{-16pt}
	\caption{Visual comparison with state-of-the-art 3D-aware GANs on SHHQ-HS.
		Our results have similar visual quality to existing 3D-aware GANs that do not handle expression and pose control.  (\textbf{Best viewed with zoom})}
	\label{fig:comparison}
\end{figure}

\begin{figure}[t!]
	\includegraphics[width=\columnwidth]{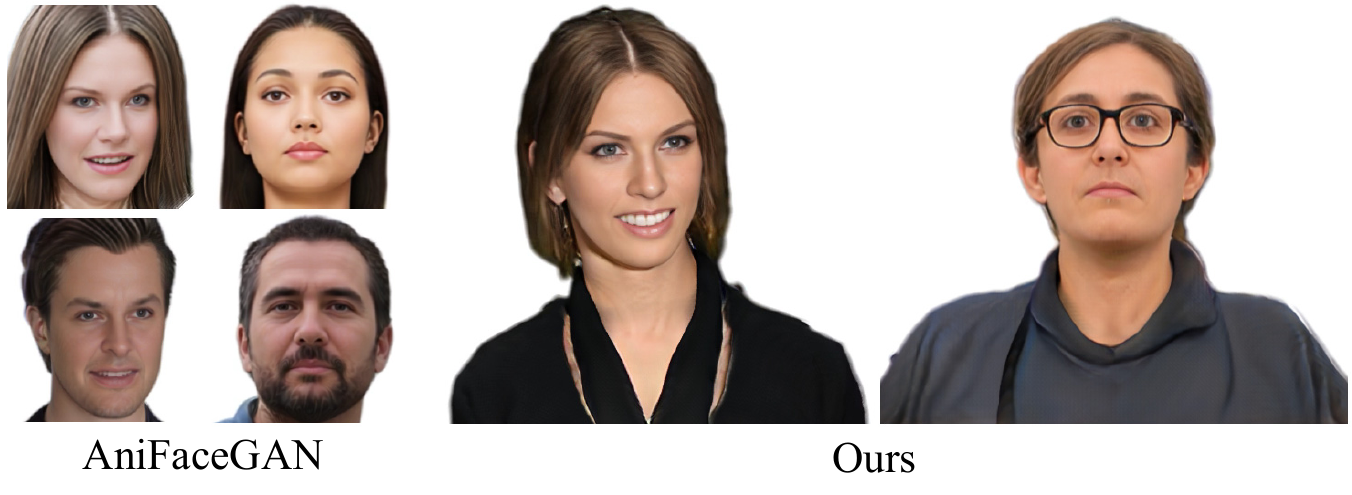}
	\vspace{-17pt}
	\caption{Visual comparison with AniFaceGAN.
	}
	\label{fig:comparison_anifacegan}
\end{figure}

Table~\ref{tab:compare_fid} shows the 
FID~\cite{heusel2017gans} and KID~\cite{binkowski2018demystifying} metrics 
evaluated on both the full portrait images and the face regions. Our method has comparable scores with EG3D and GRAM-HD for face and slightly lower scores on full  image. 
Note that although EG3D has lowest scores, we found that it often generates poor geometry: the portrait surfaces are sometimes nearly planar and the visual parallax is wrong when changing viewing angles (Fig.~\ref{fig:comparison}). 
Visually inspected, our image quality is comparable with EG3D and GRAM-HD and the portraits have correct geometry, as shown in Fig.~\ref{fig:comparison}. Figure~\ref{fig:comparison_anifacegan} compares the results from AniFaceGAN and our method. Clearly, our method can generate and control much larger region.

\subsection{Ablation Study}
We then conduct ablation studies to validate the effectiveness of our algorithm design. For efficiency, we quantitatively evaluate the generation quality on the $128^2$ resolution (i.e., without manifold super-resolution) and compute the FID and KID metrics using 5K generated and real images. The results are summarized in Table~\ref{tab:ablation}.

\begin{table}[t!]
	\caption{Quantitative comparison with state-of-the-art 3D-aware GANs on SHHQ-HS. Note that EG3D and GRAM-HD do not provide any expression and pose control, and AniFaceGAN only generates head images.  FID and KID ($\times 100$) are computed with 20K randomly generated images and 20K real ones. ~$^*$: Although EG3D has lowest FID and KID scores, it often generates planar geometry; see Fig.~\ref{fig:comparison}. ~$^\dagger$: AniFaceGAN is trained on $128^2$ by \cite{wu2022anifacegan} and evaluated with  $256^2$ rendering.}
	\label{tab:compare_fid}
	\vspace{-2pt}
	\begin{tabular}{lcccc}
		\toprule
		\multirow{2}{*}{Method} & \multicolumn{2}{c}{Face $256^2$} &\multicolumn{2}{c}{Full $512^2$}\\ 
		&FID$\downarrow$ &KID$\downarrow$ &FID$\downarrow$ &KID$\downarrow$\\ 
		\midrule
		EG3D & 5.63$^*$ & 0.20$^*$ & 6.81$^*$ & 0.26$^*$\\
		GRAM-HD$_{64\rightarrow512}$\!\! & 8.01\,\,\,~\!\! & 0.41\,\,\,~\!\! & 7.75\,\,\,~\!\! & 0.29\,\,\,~\!\!\\
		GRAM-HD$_{128\rightarrow512}$\!\! & 8.14\,\,\,~\!\! & 0.30\,\,\,~\!\! & 8.82\,\,\,~\!\! & 0.28\,\,\,~\!\!\\
		AniFaceGAN & \!\!11.56$^\dagger$ \, \!\!& 0.66$^\dagger$ & N/A\,\,\,~\!\! & N/A\,\,\,~\!\!\\
		Ours & 7.64\,\,\,~\!\! & 0.43\,\,\,~\!\! & \!\!10.10\,\,\,~\!\! \, \!\!& 0.43\,\,\,~\!\!\\
		\bottomrule
	\end{tabular}
\end{table}

\paragraph{Discriminator Combinations} Our full method uses three discriminators for training: $D_{whole}$, $D_{face}$, and $D_{torso}$.  Table~\ref{tab:ablation} shows that using $D_{whole}$ alone is clearly deficient, as demonstrated by the poor FID and KID scores for face. Combining $D_{whole}$ with $D_{face}$ (i.e., without $D_{torso}$) significantly improved the face quality, but the full portrait quality gets much worse than using only $D_{whole}$. The FID and KID scores of our full method are low for both the full portrait images and face regions. 

\begin{figure}[t!]
	\includegraphics[width=\columnwidth]{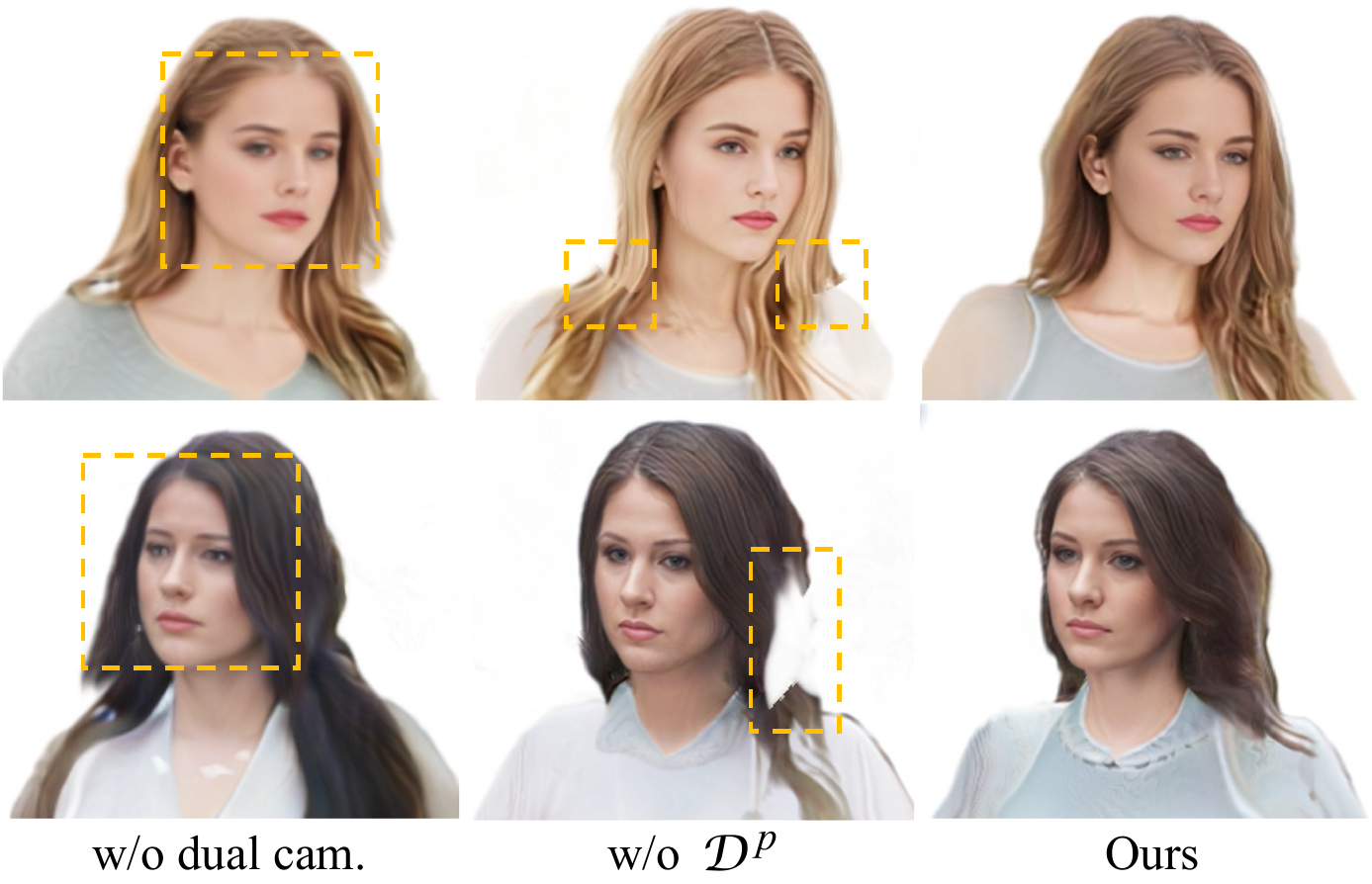}
	\vspace{-16pt}
	\caption{Ablation on the dual camera training scheme and pose deformation CNN $\mathcal{D}^p$ (trained on $128^2$ and rendered on $256^2$ ). See the suppl. video for more results.}
	\label{fig:ablation_dualcam_lbs}
\end{figure}

\paragraph{Dual-Camera Training} To validate our dual-camera rendering and adversarial learning scheme, we train a variant of our method without a separate face camera for training. For this variant, we locate faces in the rendered head-shoulder portrait images during training, and apply face discriminator on cropped and aligned faces. As we can see from Table~\ref{tab:ablation}, our method with dual cameras for training significantly outperforms such an image cropping strategy in terms of face quality with slightly lower full-image FID. Some visual examples can be found in Fig.~\ref{fig:ablation_dualcam_lbs}.

\paragraph{Pose Deformation Processing Module} 

Table~\ref{tab:ablation} (last row) shows that removing our pose deformation processing CNN $\mathcal{D}^p$, 
which degrades to a simple skinning weight assignment strategy,
leads to a quality drop for the full portrait image.  
Figure~\ref{fig:ablation_dualcam_lbs} visually compares two typical generation results. Without $\mathcal{D}^p$, sharp discontinuities will occur for long hairs under head movements, whereas our full method produces plausible results without obvious artifacts.

\begin{table}[t!]
	\caption{Ablation study on discriminator settings and the pose deformation processing CNN $\mathcal{D}^p$. $*$: w/o dual-camera training, i.e., applying $D_{face}$ on faces cropped from the rendered portrait images.  FID and KID ($\times 100$) are computed with 5K randomly generated images and 5K real ones.}
	\label{tab:ablation}
	\vspace{-2pt}
	\begin{tabular}{cccc|cccc}
		\toprule
		\multicolumn{4}{c|}{Method}&\multicolumn{2}{c}{Face $128^2$} & \multicolumn{2}{c}{Full $128^2$} \\
		\!\!\!\small $D_{whole}$\!\!\! & \!\!\!\small $D_{face}$\!\!\! & \!\!\!\small $D_{torso}$\!\!\! & \!\!\small $\mathcal{D}^p$\!\! &  FID$\downarrow$\!\! &  KID$\downarrow$\!\! &  \!FID$\downarrow$\!\! & \!KID$\downarrow$\!\! \\
		\hline
		$\checkmark$ & $\checkmark$ & $\checkmark$ & $\checkmark$ & \!\!\tikzcircle[fill={rgb,255:red,118;green,167;blue,151}]{2.5pt} 11.26\!\! & \!\!\tikzcircle[fill={rgb,255:red,118;green,167;blue,151}]{2.5pt} 0.57\!\! & \!\!\tikzcircle[fill={rgb,255:red,118;green,167;blue,151}]{2.5pt} 16.68\!\! & \!\!\tikzcircle[fill={rgb,255:red,118;green,167;blue,151}]{2.5pt} 0.97\!\! \\
		$\checkmark$ & $\times$ & $\times$ & $\checkmark$ & \!\!\tikzcircle[fill={rgb,255:red,216;green,99;blue,68}]{2.5pt} 23.00\!\! & \!\!\tikzcircle[fill={rgb,255:red,216;green,99;blue,68}]{2.5pt} 1.48\!\! & \!\!\tikzcircle[fill={rgb,255:red,234;green,194;blue,130}]{2.5pt} 18.11\!\! & \!\!\tikzcircle[fill={rgb,255:red,234;green,194;blue,130}]{2.5pt} 1.06\!\! \\
		$\checkmark$ & $\checkmark$ & $\times$ & $\checkmark$ & \!\!\tikzcircle[fill={rgb,255:red,118;green,167;blue,151}]{2.5pt} 10.68\!\! & \!\!\tikzcircle[fill={rgb,255:red,118;green,167;blue,151}]{2.5pt} 0.52\!\! &  \!\!\tikzcircle[fill={rgb,255:red,216;green,99;blue,68}]{2.5pt} 22.58\!\! & \!\!\tikzcircle[fill={rgb,255:red,216;green,99;blue,68}]{2.5pt} 1.49\!\!\\
		$\checkmark$ & $*$ &$\checkmark$ & $\checkmark$ & \!\!\tikzcircle[fill={rgb,255:red,234;green,194;blue,130}]{2.5pt} 17.89\!\! & \!\!\tikzcircle[fill={rgb,255:red,216;green,99;blue,68}]{2.5pt} 1.33\!\! & \!\!\tikzcircle[fill={rgb,255:red,118;green,167;blue,151}]{2.5pt} 14.05\!\! & \!\!\tikzcircle[fill={rgb,255:red,118;green,167;blue,151}]{2.5pt} 0.69\!\!\\
		\hline
		$\checkmark$ & $\checkmark$ & $\checkmark$ & $\times$ & \!\!\tikzcircle[fill={rgb,255:red,118;green,167;blue,151}]{2.5pt} 10.88\!\! & \!\!\tikzcircle[fill={rgb,255:red,118;green,167;blue,151}]{2.5pt} 0.52\!\! & \!\!\tikzcircle[fill={rgb,255:red,234;green,194;blue,130}]{2.5pt} 19.27\!\! & \!\!\tikzcircle[fill={rgb,255:red,234;green,194;blue,130}]{2.5pt} 1.23\!\!\\
		\bottomrule
	\end{tabular}
\end{table}

\subsection{Talk Video Generation}
\begin{figure}[t!]
	\includegraphics[width=\columnwidth]{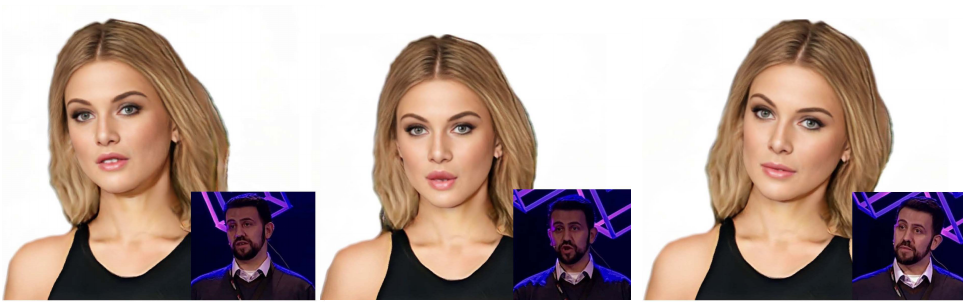}
	\vspace{-16pt}
	\caption{Talk video generation driven by real person.}
	\label{fig:driving}
\end{figure}

We further test our trained model on the task of generating  videos of talking portraits driven by real videos. Specifically, we selected some talk videos from the 300-VW dataset~\cite{shen2015first} and track the 3DMM expression and SMPL head-shoulder pose using the methods of \cite{deng2019accurate} and \cite{joo2021exemplar}, respectively. Simple temporal smoothing is applied on the estimation results. Then we transfer the tracked results to our generated human characters to obtain virtual talk videos. Figure~\ref{fig:driving} shows some typical examples of our results where a generated virtual character moves following the real person. 
See the supplementary video for continuous animations of our results.

\section{Conclusion}
We have presented a novel 3D-aware GAN for animatable head-shoulder portrait generation, a new task not addressed by previous methods. We identified several key issues when extending existing techniques to this new task and proposed targeted algorithms to tackle them. 
We demonstrate that by training a corpus of unstructured 2D images,  
our method can generate diverse and high-quality 3D portraits with 
controllable facial expression as well as head and shoulder movements. 
We believe our work represents one step forward towards auto-creating video avatars for real-world applications. 

\begin{figure}[t!]
	\includegraphics[width=\columnwidth]{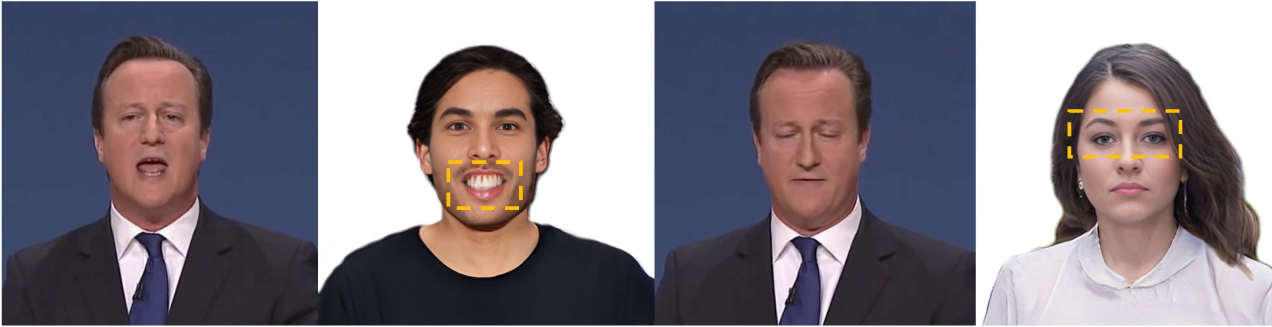}
	\vspace{-16pt}
	\caption{Limitations on more extreme expressions and closed eyes. For each image pair, the left one is the reference image and the right one is the animated result. }
	\label{fig:limitation}
\end{figure}

\paragraph{Limitations} 
Our method still has several limitations. 
It may produce artifacts under human poses and expressions that are not present in training data distribution, as shown in Fig.~\ref{fig:limitation}. 
In fact, the facial expression variation in SHHQ-HS is rather limited, lacking images with extreme expression and closed eyes. The visual quality of the inner mouth region (e.g., teeth) is not satisfactory, which is also partially due to limited data samples.
Additionally, our current method lacks the ability to control other attributes, such as eye gaze and environment lighting. 
We plan to further explore and address these issues in our future research.

\paragraph{Ethics and responsible AI considerations}
This work aims to design an animatable 3D-aware human portrait generation method for the application of virtual avatars. It is not intended to create content that is used to mislead or deceive. However, it could still potentially be misused. We condemn any behaviors of creating misleading or harmful contents and are interested in applying this technology for advanced forgery detection. Currently, the images generated by this method contain visual artifacts that can be easily identified. The method's performance is affected by the biases in the training data. One should be careful about the data collection process and ensure unbiased distributions of race, gender, age, among others.  

{\small
	\bibliographystyle{ieee_fullname}
	\bibliography{bib}
}

\clearpage

\appendix

\begin{figure*}[t!]
	\centering
	\includegraphics[width=0.99\textwidth]{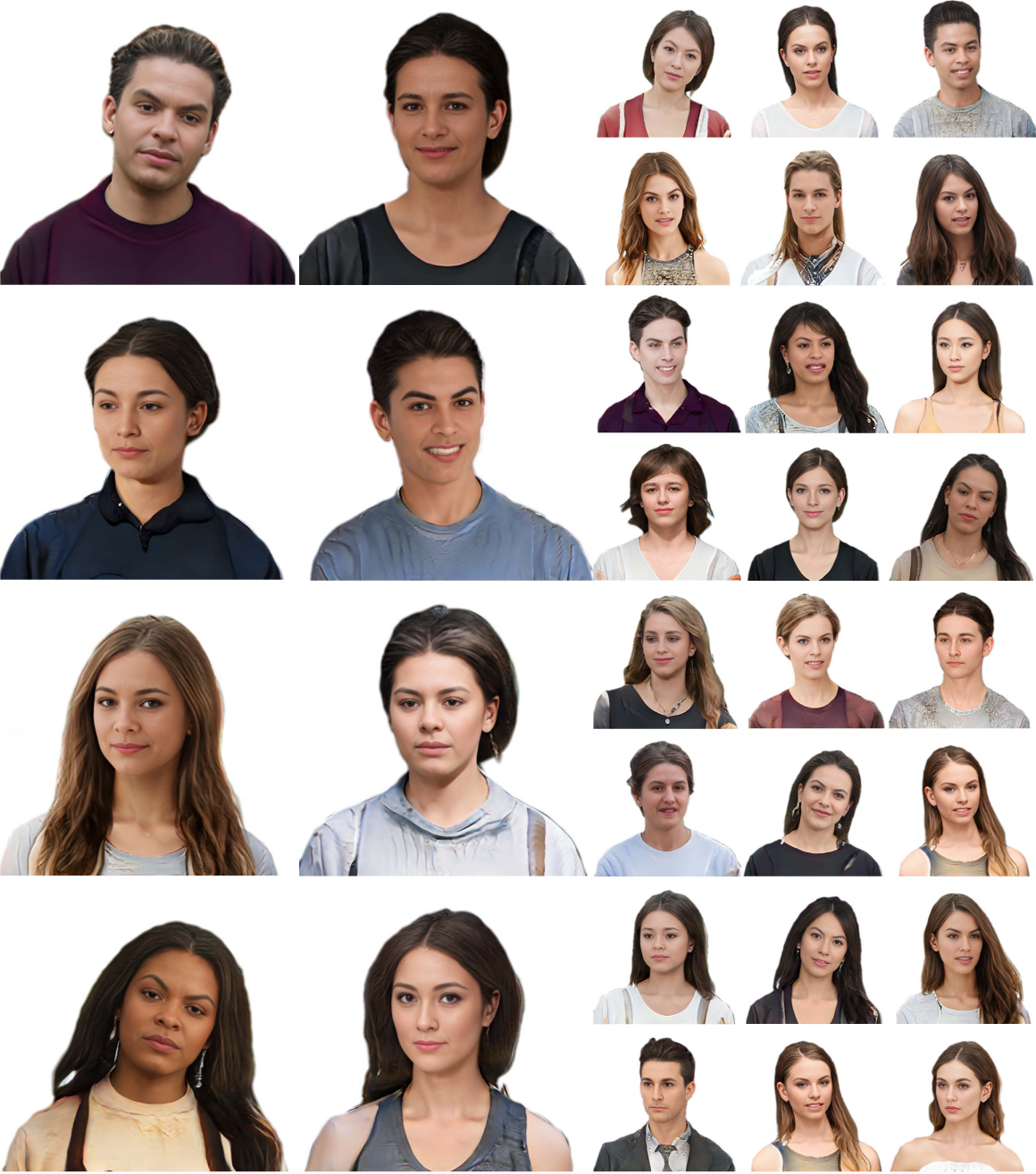}
	\caption{Uncurated portrait generation results from our method.}
	\label{fig:results_uncurated}
\end{figure*}

\begin{figure*}[t!]
	\includegraphics[width=0.98\textwidth]{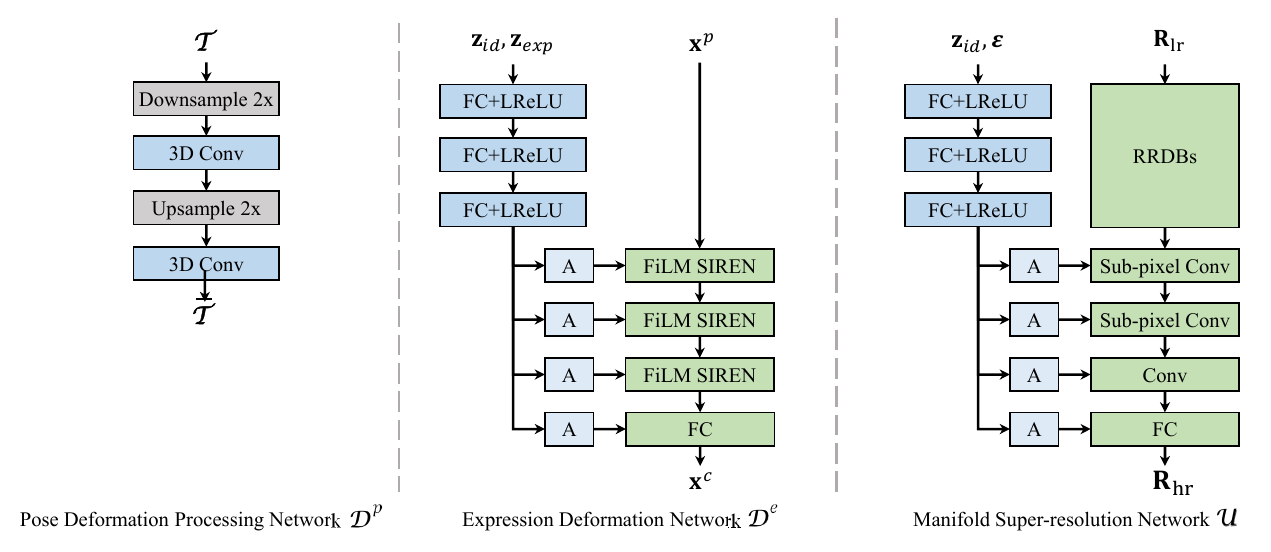}
	\caption{Network structures. $\mathcal{D}^p$ has two 3D conv layers with kernel size $9\times 9\times 5$. Both $\mathcal{D}^e$ and $\mathcal{U}$ have a StyleGAN-like structure \cite{karras2019style} with a mapping MLP network and a backbone. The backbone of $\mathcal{D}^e$ is mainly based on 3 FiLM SIREN layers~\cite{chan2021pi} while  $\mathcal{U}$ uses 4 Residual-in-Residual Dense Blocks (RRDBs) \cite{wang2018esrgan} and 2 sub-pixel conv layers \cite{shi2016real}.}
	\label{fig:network_structure}
\end{figure*}
	
\end{document}